\documentclass[a4paper]{article}

\usepackage[hyperfootnotes=false]{hyperref}
\usepackage{INTERSPEECH2019}

\usepackage{graphicx}
\graphicspath{{fig/}}
\newcommand{\imagesep}{\vspace*{-6pt}}
\newcommand{\tablesep}{\vspace*{-6pt}}

\usepackage{booktabs}
\usepackage{tabularx}
\usepackage{multirow}
\newcommand{\mytable}{
    \centering
    \renewcommand{\arraystretch}{1.1}
    }

\newcolumntype{C}{>{\centering\arraybackslash}X}
\newcolumntype{L}{>{\raggedright\arraybackslash}X}
\newcolumntype{R}{>{\raggedleft\arraybackslash}X}
\newcolumntype{P}[1]{>{\raggedright\arraybackslash}p{#1}}

\usepackage{amsmath}
\usepackage{amssymb}
\renewcommand{\vec}[1]{\boldsymbol{\mathbf{#1}}}

\usepackage{calc}
\usepackage{microtype}
\usepackage{url}
\usepackage{xcolor}

\newcommand\blfootnote[1]{\begingroup
                          \renewcommand\thefootnote{}\footnote{#1}
                          \addtocounter{footnote}{-1}
                          \endgroup}

\usepackage{cite}
\title{Unsupervised acoustic unit discovery for speech synthesis \\ using discrete latent-variable neural networks}
\name{Ryan Eloff$^*$, Andr\'{e} Nortje$^*$, Benjamin van Niekerk$^*$, Avashna Govender$^\dagger$, Leanne Nortje, Arnu~Pretorius, Elan van Biljon, Ewald van der Westhuizen, Lisa van Staden, Herman Kamper$^*$\vspace*{-3pt}}
\address{Stellenbosch University, South Africa \& $^\dagger$The University of Edinburgh, UK}
\email{kamperh@sun.ac.za}

\usepackage[prependcaption,textsize=scriptsize]{todonotes}
\definecolor{mycolor}{HTML}{FF6600}

\begin{document}

\maketitle

\begin{abstract}
For our submission to the ZeroSpeech 2019 challenge, we apply discrete latent-variable neural networks to unlabelled speech and use the discovered units for speech synthesis. Unsupervised discrete subword modelling could be useful for studies of phonetic category learning in infants or in low-resource speech technology requiring symbolic input. We use an autoencoder~(AE) architecture with intermediate discretisation. We decouple acoustic unit discovery from speaker modelling by conditioning the AE's decoder on the training speaker identity. At test time, unit discovery is performed on speech from an unseen speaker, followed by unit decoding conditioned on a known target speaker to obtain reconstructed filterbanks. This output is fed to a neural vocoder to synthesise speech in the target speaker's voice. For discretisation, categorical variational autoencoders (CatVAEs), vector-quantised VAEs (VQ-VAEs) and straight-through estimation are compared at different compression levels on two languages. Our final model uses convolutional encoding, VQ-VAE discretisation, deconvolutional decoding and an FFTNet vocoder.
We show that decoupled speaker conditioning intrinsically improves discrete acoustic representations, yielding competitive synthesis quality compared to the challenge baseline.
\blfootnote{\hspace{-4.33331pt}$^*$These authors contributed equally. The other authors contributed to this work during a two-week coding sprint at Stellenbosch University.}
\end{abstract}

\section{Introduction}

For many languages, it is difficult or impossible to collect the annotated resources required for training supervised automatic speech recognition (ASR) models~\cite{besacier+etal_speechcom14}.
Zero-resource speech processing aims to develop methods that can learn directly from unlabelled speech audio.
A central problem is finding frame-level feature representations that capture meaningful linguistic contrasts, such as phonetic categories, while being invariant to factors such as a speaker's identity, gender or accent~\cite{zhang+glass_asru09,jansen+etal_icassp13,jansen+etal_icassp13b,hsu+etal_nips17}.
As part of the 2015 and 2017 Zero Resource Speech Challenges~(ZRSC), several unsupervised methods were proposed for learning \textit{continuous} acoustic representations~\cite{renshaw+etal_interspeech15,thiolliere+etal_interspeech15,versteegh+etal_sltu16,dunbar+etal_asru17,heck+etal_asru17,tsuchiya+etal_icassp18}. 
Here we consider \textit{discrete} representation learning, also referred to as `acoustic unit discovery'~\cite{lee+etal_tacl15,ondel+etal_pcs16} or `discrete subword learning'~\cite{badino+etal_interspeech15}. 
From a scientific perspective, discrete representations could be useful in cognitive models to study phonetic category learning in human infants~\cite{feldman+etal_ccss09,rasanen_speechcom12,schatz+feldman_ccn18,shain+elsner_naacl19}.
From a technology perspective, such features could be used in downstream speech applications requiring symbolic or sparse input, e.g., for faster retrieval in speech search systems~\cite{levin+etal_icassp15,settle+etal_interspeech17}.
Here we consider the downstream task of speech synthesis within the context of the ZRSC'19.

Specifically, the task is to perform acoustic unit discovery on an unlabelled speech collection, to build a synthesiser in a target speaker's voice using the discovered units, and to then synthesise speech from unseen speakers in the target speaker's voice.
This is related to voice conversion~\cite{kain+etal_icassp98,chou+etal_interspeech18}, where speech from one speaker needs to be synthesised in another speaker's voice. 
But, importantly, the task here involves unit discovery which compresses spoken input~\cite{muthukumar+black_icassp14,scharenborg+etal_icassp18}. 
The trade-off between the level of compression and intelligibility is explicitly evaluated. 

For unit discovery, we use an
autoencoder (AE) neural network architecture, which compresses its input and then reconstructs it from a latent layer.
AEs are well-suited to our task 
since compression can be directly controlled through the latent layer dimensionality~\cite{zeiler+etal_icassp13,badino+etal_icassp14,badino+etal_interspeech15}.
Unfortunately, naive discretisation within AEs 
is not possible since gradient updates cannot be calculated directly. 
We build on a number of recent stochastic methods which allows these gradients to be estimated~\cite{bengio+etal_arxiv13,raiko+etal_iclr15,toderici+etal_iclr16,jang+etal_iclr17,maddison+etal_iclr17,vandenoord+etal_neurips17}.

Concretely, we investigate different unsupervised discrete latent-variable neural networks for acoustic unit discovery and subsequent speech synthesis.
Our AE architecture takes Mel-frequency cepstral coefficients (MFCCs) as input, performs intermediate discretisation to find discrete units, and produces log-Mel filterbank features as output. 
These are fed to an autoregressive neural vocoder, producing synthesised speech.
During evaluation, the goal is to synthesise speech in a target speaker's voice given input from an unseen speaker.
In order to separate subword from speaker modelling, we feed a speaker's identity to the AE's decoder module during training.
Speaker conditioning is not used in the AE's encoder and discretisation steps, so discrete units can be obtained for an arbitrary speaker.
At test time, units are obtained for an unseen test speaker; these units are then decoded  by conditioning on the target speaker's voice using the combination of the AE's decoder and the vocoder (trained on the target speaker's voice).
In contrast to the feed-forward AEs of~\cite{zeiler+etal_icassp13,badino+etal_icassp14,badino+etal_interspeech15}, we incorporate temporal context by using a convolutional encoder and deconvolutional decoder in our~AE.

A similar model using a vector-quantised variational autoencoder (VQ-VAE) 
was recently proposed in~\cite{chorowski+etal_arxiv19}.
Our approach differs from this model in several ways, 
corresponding to the main contributions of our work.
Firstly, we do not train our model end-to-end, but rather train separate compression and vocoding models. 
This is not necessarily superior, but it has benefits. E.g., simpler/more complex vocoders could be used with the same compression model depending on computational resources.
Secondly, we specifically consider the impact of decoupling unit discovery from speaker modelling. 
We show that this produces better discrete latent features in an intrinsic cross-speaker evaluation task.
Thirdly, we consider different discretisation methods; we show that the categorical VAE (CatVAE)~\cite{jang+etal_iclr17,maddison+etal_iclr17} and straight-through estimation (STE)~\cite{bengio+etal_arxiv13,raiko+etal_iclr15,toderici+etal_iclr16} both perform competitively to the VQ-VAE.
Finally, we evaluate speech synthesis performance using human evaluations as part of ZRSC'19 on both English and Indonesian data. 

\section{Unsupervised discrete representation learning for speech synthesis}

A corpus of unlabelled speech from multiple speakers in a single language is used for training our discrete latent-variable neural networks. 
A known training-set speaker is specified as the target voice in which to synthesise speech.
At test time, the proposed system is provided with new utterances from unseen speakers.
For each test utterance, the system performs unit discovery on the input, producing a transcription-like encoding using the learned 
discrete symbols.
The test utterance is then re-synthesised in the target speaker's voice based on these symbols~\cite{dunbar+etal_interspeech19}.

\begin{figure}[!t]
    \centering
    \includegraphics[scale=0.75]{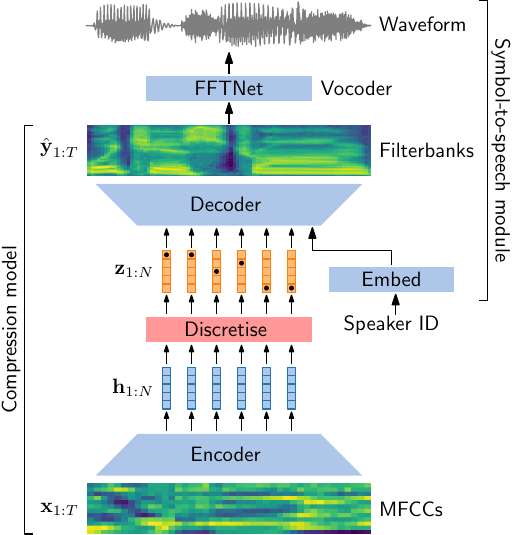}
    \vspace{2pt}
    \imagesep
    \caption{Discrete latent-variable neural networks are used for unit discovery and speech synthesis.
    At training time, the speaker ID matches the input. At test time it is set to a target speaker. The compression model and vocoder is trained separately.} 
    \label{fig:model}
    \vspace*{-10pt}
\end{figure}

Our specific approach is shown in Figure~\ref{fig:model}.
The system~takes MFCCs with deltas and double-deltas as input, 
$\vec{x}_{1:T} = \vec{x}_1, \vec{x}_2, \ldots, \vec{x}_T$,
with each $\vec{x}_t\in\mathbb{R}^{39}$.
These are encoded into latent continuous representations $\vec{h}_{1:N} $, which are discretised into symbolic representations $\vec{z}_{1:N}$.
The sequence of symbols $\vec{z}_{1:N}$ preserves time-information at a fixed rate, with $N$ depending
on the length of the input $T$ and the amount of downsampling in the encoder (\S\ref{sec:architectures}).
A symbol $\vec{z}$ takes on a different form depending on the discretisation method: it can be binarised (STE), one-hot (CatVAE), or an embedding selected from a codebook (VQ-VAE).
But, importantly, the input is encoded into a finite set of symbols.
These symbols are decoded to 45-dimensional filterbanks $\vec{y}_{1:N}$ using a decoder module. Since the encoder input and the decoder output is obtained from the same original waveform, we refer to this compression model as an autoencoder~(AE). 

At training time, the AE's decoder is conditioned on the input speaker identity.
This means that, in principle, the encoder does not need to preserve speaker-specific information, 
thereby decoupling acoustic unit discovery from speaker modelling.
For this same reason we use MFCCs, which tend to be speaker-independent~\cite{davis+mermelstein_tassp80}, as input, while fitlerbanks, which retain speaker-specific properties, are used as intermediate output.
We also experimented with filterbank input, but MFCCs worked better.
The encoder and discretiser are speaker-independent and can therefore be applied to unseen speakers at test time.
The decoder is then conditioned on the target speaker and its output is provided to an FFTNet vocoder~\cite{jin+etal_icassp18}, trained only on the target speaker, to obtain synthesised speech in the desired voice.
Voice conversion 
therefore occurs in the AE's decoder, not the~vocoder.

The whole system can be trained without parallel data.
In contrast to the end-to-end methodology of~\cite{chorowski+etal_arxiv19}, we train the compression model 
separately from the vocoder.
The combination of the AE's decoder with the vocoder could thus be described as a symbol-to-speech module, as shown in Figure~\ref{fig:model}.

Building on recent advances in discrete representations in neural networks, we use three discretisation strategies to convert the encoder output $\vec{h}$ to a  discrete latent representation $\vec{z}$.\footnote{Here, $\vec{h}$ refers to a single latent vector in the sequence $\vec{h}_{1:N}$. The scalar $h_k$ is the $k^\textrm{th}$ dimension of $\vec{h}$. The same convention is used for $\vec{z}$.}

%

\subsection{Straight-through estimation (STE) binarisation}
\label{sec:ste}

A direct way to discretise would be to binarise the continuous encoder output vectors.
If $\vec{h}$ is a $K$-dimensional continuous vector obtained through a $\tanh$ activation, i.e.\ $\vec{h} \in [-1, 1]^K$, each dimension of $\vec{z}$ could be set so that $z_{k} = 1$ if $h_{k} \geq 0$ or $z_{k} = -1$ otherwise.
The result would be a binarised vector $\vec{z} \in \{-1, 1\}^K$.
The gradients required for back-propagating through such a thresholding operation cannot be directly derived, so we use the straight-through estimation (STE) method~\cite{raiko+etal_iclr15,toderici+etal_iclr16}. During training, binarisation is performed stochastically with $z_{k} = h_{k} + \epsilon$, where $\epsilon$ is sampled noise:
\vspace*{-5pt}
\begin{equation*}
    \epsilon =
    \begin{cases}
        1 - h_k & \text{with probability } \frac{1 + h_k}{2} \\
        -h_k - 1 & \text{with probability } \frac{1 - h_k}{2}  \\
    \end{cases}
\vspace*{-5pt}
\end{equation*}
Since $\epsilon$ is zero-mean,
the derivative of the expected value of $z_k$ is $\frac{\partial \mathbb{E}[z_k]}{\partial h_k} = 1$, 
so gradients are passed unchanged through the binarisation layer: $\frac{\partial L}{\partial \vec{h}} \approx \frac{\partial L}{\partial \vec{z}}$.
We use a squared loss~$L$. 

\subsection{Vector-quantised variational autoencoder (VQ-VAE)}
\label{sec:vqvae}

The vector-quantised variational autoencoder (VQ-VAE)~\cite{vandenoord+etal_neurips17} maintains a codebook of prototype embedding vectors $\{ \vec{e}_k \}_{k = 1}^K$.
During inference, the encoder output $\vec{h}$ is used to select the closest codebook embedding to serve as the quantised representation, i.e., $\vec{z} = \vec{e}_k$ where $k = \text{argmin}_j || \vec{h} - \vec{e}_j ||^2$.
Once again, gradients cannot be derived for this operation, so we approximate it as $\frac{\partial L}{\partial \vec{h}} \approx \frac{\partial L}{\partial \vec{z}}$.
VQ-VAE discretisation uses a loss $L$ composed of three terms.
The first is the negative log likelihood, which reduces to a reconstruction loss when assuming a spherical Gaussian output distribution: 
$-\log p_{\vec{\theta}}(\vec{y}_{1:T} | \vec{z}_{1:N}) = - \sum_{t = 1}^T \left( c - \frac{1}{2\sigma^2} ||\vec{y}_t - \hat{\vec{y}}_t ||^2 \right)$, 
with $\vec{\theta}$ denoting the decoder parameters, $\vec{y}_t$ the target filterbanks, and $\hat{\vec{y}}_t$ the predicted decoder output.
The second loss term, $||\text{sg}(\vec{h}) - \vec{z}||^2$, updates the embeddings in the codebook, with $\text{sg}(\cdot)$ denoting the stop-gradient operation.
Third, a `commitment loss' encourages the encoder output $\vec{h}$ to lie close to a codebook embedding: $\beta ||\vec{h} - \text{sg}(\vec{z})||^2$.
The three terms can be weighed by changing the hyperparameters $\sigma$ and $\beta$.
We use $\sigma = 10^{-6}$ and $\beta = 25$.

\subsection{Categorical variational autoencoder (CatVAE)}
\label{sec:catvae}

As in a standard VAE~\cite{kingma+welling_arxiv13}, the categorical VAE (CatVAE)~\cite{jang+etal_iclr17,maddison+etal_iclr17} optimises a lower bound for $\log p_{\vec{\theta}}(\vec{y}_{1:T})$ using an encoder $q_{\vec{\phi}}(\vec{z}|\vec{x}_{1:T})$ to approximate the intractable posterior $p_{\vec{\theta}}(\vec{z}|\vec{x}_{1:T})$, with $\vec{\phi}$ and $\vec{\theta}$ denoting the encoder and decoder weights, respectively. 
In a standard VAE, $\vec{z}$ is drawn from a multivariate Gaussian using the reparametrisation trick. 
For the CatVAE, $\vec{z}$ is a one-hot vector sampled from a categorical distribution, which does not have a direct reparametrised form.
The Gumbel-softmax distribution approximates the categorical distribution and can be reparametrised~\cite{jang+etal_iclr17}.
It is defined as $P(z_k) = \frac{ \exp\left\{ (\log \pi_k + g_k) / \tau \right\} }{ \sum_{k = 1}^K \exp\left\{ (\log \pi_k + g_k) / \tau \right\} }$ 
where $g_1, g_2, \ldots, g_K$ are samples from a $\textrm{Gumbel}(0, 1)$ distribution and $\tau$ is a temperature parameter.
Sampling from this distribution with small $\tau$ resembles 
a $K$-component categorical distribution with weights $\vec{\pi} = \{\pi_k\}_{k = 1}^K$.
Our AE encoder outputs these weights $\vec{h} = \log\vec{\pi}$. 

The model is trained using the evidence lower bound for $\log p_{\vec{\theta}}(\vec{y}_{1:T})$, which reduces to a reconstruction term (the same as for the VQ-VAE) and a Kullback-Leibler (KL) regularisation term.
This KL term encourages the latent features $\vec{z}$ to be similar to a prior $p(\vec{z})$, chosen as the uniform categorical distribution.
The term can thus be optimised directly since it is the KL divergence between two categorical distributions: the one predicted for $\vec{z}$, with mass $\vec{\pi}$ given by the encoder, and the uniform prior. 
During training, we anneal $\tau$ linearly from 1 to 0.1.

\vspace*{-3pt}
\subsection{Neural network architectures}
\vspace*{-1pt}
\label{sec:architectures}

Each of the above methods are used as the discretisation layer in Figure~\ref{fig:model}.
Our goal is to compress input speech into a discrete set of symbols which can be used for speech synthesis.
The degree of compression (the bitrate) can be controlled by changing the number of unique symbols, set through $K$.
To further reduce bitrate, the encoder architecture can downsample the input, producing \mbox{discrete features at a fixed but lower rate than the input.}

Specifically, all our models use a first pre-processing convolutional layer without downsampling.
This is followed by a convolutional layer with a stride of 2, producing output at half the rate of its input.
This strided convolutional layer can be repeated, each downsampling by a factor of 2.
If not specified, we use two such layers, i.e.\ for MFCC input with $T$ frames, the model produces $N = T/4$ discrete vectors.
For decoding, transposed convolutions mirror the convolutions in the encoder. A linear output layer produces the $T$-frame filterbank output.
We also experimented with recurrent architectures, but the convolutional approach proved superior and was faster to train.

Based on development experiments on English data (\S\ref{sec:setup}), we chose the same architecture for VQ-VAE and CatVAE discretisation, but a slightly different architecture for STE.
For the STE model, we use convolutional gated recurrent units (ConvGRU)~\cite{toderici+etal_cvpr17} to downsample.
A ConvGRU is a type of recurrent cell which  applies convolutions over its input. 
Rather than straight-forward transposed convolutions, our STE model also uses the more efficient pixel shuffling operation~\cite{shi+etal_cvpr16}.
The VQ-VAE/CatVAE models use residual convolutions with batch normalisation and standard transposed convolutions for decoding.
In the STE model, the decoder is conditioned on a 250-dimensional speaker embedding, while the VQ-VAE/CatVAE models use a speaker dimensionality of 128; these are trained jointly with the rest of the model.
We use Adam optimisation~\cite{kingma+ba_iclr15}.
Other architectural choices are given in our code repository.\footnote{\scriptsize \url{https://github.com/kamperh/suzerospeech2019}}

For speech synthesis, we use FFTNet as a vocoder (Figure~\ref{fig:model} top).
FFTNet is an auto-regressive neural model which generates raw audio waveforms given filterbanks as input.
We also experimented with WaveNet~\cite{vandenoord+etal_arxiv16}, which produced slighly better quality but required much longer training times and occasionally generated noise.
Our final model consists of a stack of 11 FFTNet layers with 256 channels each, giving a receptive field of 2048 samples.
We quantise the raw waveforms using $\mu$-law encoding and a softmax over 256 classes to model the distribution over the next sample.
We train FFTNet on clean filterbanks, extracted directly from the target speaker. 
We also tried to train directly on the output of the different AEs, but this performed worse.
Further work is needed to understand this observation.

\section{Experimental setup and evaluation}
\label{sec:setup}
We perform experiments on two languages. 
We compare models on English data, and then apply the best models unaltered
on the ZRSC'19 evaluation language, Indonesian, a low-resource Austronesian language widely used as a lingua franca~\cite{sakti+etal_ococosda08,sakti+etal_tcast08}.
For both languages, training data for acoustic unit discovery consists of around 15~h from 100 speakers. 
Data from a target voice is used both for acoustic discovery and for training the vocoder.
Two English and one female Indonesian target voice dataset of around 2~h each are used.
Test data consists of 30~min in both English and Indonesian from 24 and 15 speakers, respectively.

As an intrinsic measure of discriminability, 
we use the ABX task~\cite{schatz+etal_interspeech13}. 
Using a particular feature representation, ABX asks whether a triphone $X$ is more similar to triphone $A$ or triphone $B$, where $A$ and $X$ are different instances of the same triphone (e.g.\ `beg') and $B$ differs in the middle phone (e.g.\ `bag').
To explicitly measure invariance to speaker, $A$ and $B$ comes from the same speaker, while $X$ comes from another.
As a distance metric, the average frame-wise cosine distance along the dynamic time warping (DTW) alignment path is used. 
ABX is reported as an aggregated error rate over minimal pairs. 

To measure the degree of compression, 
bitrate is calculated as $\frac{M \sum_{m = 1}^M P(\vec{z}_m) \log_2 P(\vec{z}_m)}{D}$, where $M$ is the total number of symbols in the test data, $P(\vec{z}_m)$ is the estimated probability of symbol $\vec{z}_m$, and $D$ is the total duration of the data in seconds.
To measure synthesis quality, human judges transcribe the synthesised speech.
By comparing this to the ground truth, a character error rate (CER) is calculated.
Humans also rate similarity to the target voice and give a mean opinion score (MOS), both on a scale from 1 to 5 (higher is better).

The ZRSC'19 baseline system uses a pipeline of a Dirichlet process Gaussian mixture model (DPGMM) for acoustic unit discovery~\cite{ondel+etal_pcs16}, and the Merlin neural network speech synthesiser~\cite{wu+etal_ssw16} trained on the discovered units from the target speaker.
A topline system feeds output from a supervised ASR system to a supervised speech synthesiser, both trained on transcriptions.

\section{Experimental results}

\subsection{Discretisation method comparison}
\label{sec:discrete_eval}

The `ABX on latent' column in Table~\ref{tbl:abx_speaker} compares the different discretisation methods on the English test data, with ABX performed directly on the discrete symbols.
$K$ is set so that 512 unique symbols can be obtained in all models, but not all the symbols are necessarily utilised, so the bitrates differ.
The form of the latent symbols $\vec{z}$ is very different for the different methods: binarised for STE, one-hot for CatVAE, or an embedding selected from a codebook for VQ-VAE.
The cosine-DTW distance metric used in ABX might not be equally well-suited to all of these.
To make a fairer comparison, we pass the encoded symbols through the decoder for each model, and compare all the models on the reconstructed filterbank $\hat{\vec{y}}$ obtained at the output from each AE.\footnote{One way to interpret this evaluation of the symbolic features is to see the decoder as part of the ABX distance metric for each of the methods.}
All methods are therefore compared based on the same feature type.
ABX scores based on decoder outputs with and without speaker conditioning are shown in the second and third columns of Table~\ref{tbl:abx_speaker}.
In all cases, VQ-VAE performs best.
We also observe that the CatVAE's one-hot latent symbols are particularly ill-suited to direct ABX evaluation: 45.6\% error rate compared to 24.3\% and 28.7\% when using decoder outputs.

\begin{table}[!t]
    \mytable
    \caption{ABX (\%) on English for different discrete latent-variable neural models with and without speaker conditioning.} 
    \tablesep
    \begingroup
    \eightpt
    \begin{tabularx}{\linewidth}{@{}l@{\ \ }cCCr@{}}
        \toprule
        & \multicolumn{2}{c}{ABX on decoder output} \\ 
        \cmidrule{2-3} 
        Model & No spkr cond.& Spkr cond. & \vspace*{-1.7\baselineskip}ABX on latent & Bitrate\\
        \midrule
        STE & 27.5  & 26.4 & 31.5 & \hphantom{0}116 \\
        VQ-VAE & 26.0  & 22.1 & 28.6 & \hphantom{0}190 \\
        CatVAE & 28.7  & 24.3 & 45.6 & \hphantom{0}215 \\[3pt]
        Filterbanks & - & - & 29.5 & 1735 \\
        MFCCs & - & - & 22.7 & 1738 \\
        \bottomrule
    \end{tabularx}
    \endgroup
    \label{tbl:abx_speaker}
    \vspace*{-6pt}
\end{table}

\subsection{Speaker conditioning}

Our main hypothesis is that decoupling unit discovery from speaker modelling produces intrinsically better discrete representations.
A comparison of the columns with and without speaker conditioning in Table~\ref{tbl:abx_speaker} shows that, when speaker conditioning is not used, ABX is worse for all models. 
For the last two rows in the Table~\ref{tbl:abx_speaker}, ABX was performed directly on MFCCs (inputs) and filterbanks (target outputs).
Our best overall result of 22.1\% when using the VQ-VAE is better than both of these.  
This is particularly noteworthy since this model outperforms both the input and output features on which it is trained, and it does so while performing intermediate discretisation which encodes the input at a much lower bitrate.
Discretisation on its own (ABX of 26.0\% for VQ-VAE) already leads to improvements over filterbanks (29.5\%), but it is the combination of discretisation and decoupled speaker conditioning which leads to the best score~(22.1\%).



\begin{figure}[t]
    \centering
    \includegraphics[width=0.975\linewidth]{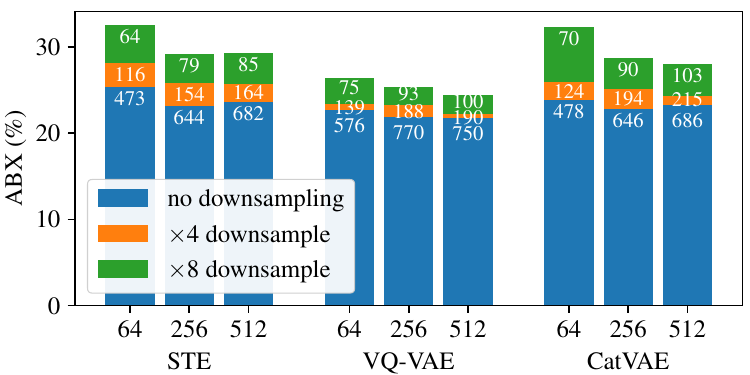}
    \imagesep
    \caption{ABX on English at different downsampling levels and numbers of symbols ($x$-axis labels). Models are evaluated on their decoder outputs.  Bitrates are shown in white on the bars.}
    \label{fig:compression_symbols}
    \vspace*{-10pt}
\end{figure}

\subsection{Results across different bitrates}

As noted in \S\ref{sec:architectures}, there are two ways to control the level of compression: $K$ can be varied, which sets the number of unique symbols, or strided convolutions in the model's encoder can be used to perform downsampling.
Figure~\ref{fig:compression_symbols} shows ABX scores on English as we vary the number of symbols (64, 256, 512) and the downsampling factor ($\times$1, $\times$4 and $\times$8) for the three discretisation methods. 
The general trend is that ABX improves when using more symbols and/or less downsampling, which is expected since this results in higher bitrates.
Another trend  is that it is typically more beneficial to use downsampling to reduce the bitrate rather than reducing the symbols.
E.g., the difference between the 256-  and 512-symbol STE bars are smaller than the differences within each bar (when downsampling is changed).

\subsection{Final results and speech synthesis}

Based on the development experiments on English, we submitted two versions of the VQ-VAE model for official evaluation to the ZRSC'19: the one uses $\times$4 and the other $\times$8 downsampling.
We did this to measure the impact of bitrate on speech synthesis quality.
Results on the English and Indonesian test data are reported in Table~\ref{tbl:eval}, 
together with the DPGMM-Merlin baseline and the supervised top-line.
Our approach performs worse on MOS and similarity compared to DPGMM-Merlin. 
On CER, however, we perform similarly or better in all cases. In terms of ABX directly on the discrete latent features, both VQ-VAE-x8 and VQ-VAE-x4 outperform the baseline---on English the latter even outperforms the supervised topline system (27.6\% compared to 29.9\% in ABX). 

We also submitted reconstructed filterbanks $\hat{\vec{y}}$ 
as auxiliary features to ZRSC'19, with the results reported in the `output' column in Table~\ref{tbl:eval}.
As in \S\ref{sec:discrete_eval}, these features give better scores than when evaluating on the symbols directly.
On English, output features from both our models outperform the supervised topline system while on Indonesian the VQ-VAE-x4 model does.
These results suggests that the main bottleneck in our overall approach is not in the AE compression model (Figure~\ref{fig:model}), but in the vocoder.
We use FFTNet mainly for computational reasons.
Although not conclusive, a comparison of our English MOS ($\sim$2.3) to the MOS for FFTNet reported in~\cite{jin+etal_icassp18} ($\sim$3.2) might indicate that we are reaching a performance ceiling with this vocoder.

\begin{table}[!t]
    \mytable
    \caption{Human and machine ZRSC'19 evaluation on the English and Indonesian test data. Higher MOS and similarity scores are better, while lower is better for the other metrics.}
    \tablesep
    \begingroup
    \eightpt
    \begin{tabularx}{\linewidth}{@{}l@{\ \ }C@{\ \ }C@{\ \ }c@{\ \ }C@{\ }C@{\ \ }r@{}}
        \toprule
        & CER & MOS & Similarity & \multicolumn{2}{@{\ \ }c}{\underline{\ \ \ ABX (\%)\ \ \ }} & \\
        Model & (\%) & [1, 5] & [1, 5] & latent & output & Bitrate \\
        \midrule
        \underline{\textit{English:}} \\[2pt]
        DPGMM-Merlin & 75 & 2.50 & 2.97 & 35.6& - & \hphantom{0}72 \\
        VQ-VAE-x8 & 75 & 2.31 & 2.49 & 30.8 & 25.1 & \hphantom{0}88\\
        VQ-VAE-x4 & 67 & 2.18 & 2.51 & 27.6 & 23.0 & 173 \\[2pt]
        Supervised & 44 & 2.77 & 2.99 & 29.9 & - & \hphantom{0}38\\        
        \midrule
        \underline{\textit{Indonesian:}} \\[2pt]
        DPGMM-Merlin & 62 & 2.07 & 3.41 & 27.5 & - & \hphantom{0}75 \\
        VQ-VAE-x8 & 58 & 1.94 & 1.95 & 26.5 & 17.6 & \hphantom{0}69 \\
        VQ-VAE-x4 & 60 & 1.96 & 1.76 & 19.8 & 14.5 &  140 \\[2pt]
        Supervised & 28 & 3.92 & 3.95 & 16.1 & - & \hphantom{0}35 \\
        \bottomrule
    \end{tabularx}
    \endgroup
    \label{tbl:eval}
    \vspace*{-10pt}
\end{table}

\vspace*{-1pt}
\section{Conclusions and future work}

We have proposed and evaluated different discrete latent-variable neural network models for unsupervised acoustic unit discovery and subsequent speech synthesis.\blfootnote{We thank Herman A.\ Engelbrecht for support and NVIDIA Corporation for sponsoring a Titan Xp GPU for this work. AG is supported by the EU's H2020 research and innovation programme under the MSCA GA 67532 (ENRICH). HK is supported by a Google Faculty Award.}
A convolutional encoder is used and acoustic unit discovery is decoupled from speaker modelling by conditioning a deconvolutional decoder on the training speaker.
At test time, the decoder is conditioned on a target speaker.
In terms of intelligibility as measured by human character error rate, our models perform similarly or better than the ZRSC'19 baseline system.
In an intrinsic evaluation of feature discriminability and speaker invariance, our discrete symbolic representation outperform the baseline by around 50\% relative, even outperforming 
a supervised topline system.
Experiments indicate that synthesis quality is impaired most by the neural vocoder, which is fed with reconstructed filterbanks from our decoder. 
In future work, we plan to consider more complex neural vocoders such as WaveNet~\cite{vandenoord+etal_arxiv16}, and to compare our approach to a fully end-to-end model.

\newpage
\bibliography{mybib}

\end{document}